\documentclass[conference]{IEEEtran}
\IEEEoverridecommandlockouts
\usepackage{cite}
\usepackage{amsmath,amssymb,amsfonts}
\usepackage{algorithmic}
\usepackage{graphicx}
\usepackage{textcomp}
\usepackage{booktabs} 
\usepackage{amsmath}

\usepackage{bm}
\usepackage{color}
\usepackage{multirow}
\usepackage[table]{xcolor}
\def\BibTeX{{\rm B\kern-.05em{\sc i\kern-.025em b}\kern-.08em
    T\kern-.1667em\lower.7ex\hbox{E}\kern-.125emX}}
\begin{document}

\title{A Review of Human-Object Interaction Detection\\
\thanks{This work has been submitted to the IEEE for possible publication. Copyright may be transferred without notice, after which this version may no longer be accessible. \\
* Corresponding author: Zhenao Wei.}
}

\author{\IEEEauthorblockN{1st Yuxiao Wang}
\IEEEauthorblockA{
% \textit{School of Furture Technology} \\
\textit{South China University of Technology}\\
% GuangZhou, China \\
ftwangyuxiao@mail.scut.edu.cn}
\and
% \IEEEauthorblockN{Qiwei Xiong}
% \IEEEauthorblockA{\textit{School of Furture Technology} \\
% \textit{South China University of Technology}\\
% GuangZhou, China \\
% xiongqiwei@vip.qq.com}
% \and
\IEEEauthorblockN{2nd Yu Lei}
\IEEEauthorblockA{
% \textit{School of Information Science \& Technology} \\
\textit{Southwest Jiaotong University}\\
% ChengDu, China \\
leiyu1117@my.swjtu.edu.cn}
\and
\IEEEauthorblockN{3rd Li Cui}
\IEEEauthorblockA{
% \textit{Group Propaganda Department} \\
\textit{Langfang Open University}\\
% LangFang, China \\
463485938@qq.com}
\and
\IEEEauthorblockN{4th Weiying Xue}
\IEEEauthorblockA{
% \textit{School of Furture Technology} \\
\textit{South China University of Technology}\\
% GuangZhou, China \\
202320163283@mail.scut.edu.cn}
\and
\IEEEauthorblockN{5th Qi Liu}
\IEEEauthorblockA{
% \textit{School of Furture Technology} \\
\textit{South China University of Technology}\\
% GuangZhou, China \\
drliuqi@scut.edu.cn}
\and
\IEEEauthorblockN{6th Zhenao Wei*}
\IEEEauthorblockA{
% \textit{School of Furture Technology} \\
\textit{South China University of Technology}\\
% GuangZhou, China \\
wza@scut.edu.cn}
% \and

}

\maketitle

\begin{abstract}
Human-object interaction (HOI) detection plays a key role in high-level visual understanding, facilitating a deep comprehension of human activities. Specifically, HOI detection aims to locate the humans and objects involved in interactions within images or videos and classify the specific interactions between them. The success of this task is influenced by several key factors, including the accurate localization of human and object instances, as well as the correct classification of object categories and interaction relationships. This paper systematically summarizes and discusses the recent work in image-based HOI detection. First, the mainstream datasets
% and evaluation metrics 
involved in HOI relationship detection are introduced. Furthermore, starting with two-stage methods and end-to-end one-stage detection approaches, this paper comprehensively discusses the current developments in image-based HOI detection, analyzing the strengths and weaknesses of these two methods. Additionally, the advancements of zero-shot learning, weakly supervised learning, and the application of large-scale language models in HOI detection are discussed. 
Finally, the current challenges in HOI detection are outlined, and potential research directions and future trends are explored.

\end{abstract}


\section{Introduction}
With the explosive growth of image data, understanding and analyzing the content within these images has become a crucial challenge. Relying solely on human vision for recognition is far from sufficient to meet the needs of modern society. Consequently, human-object interaction (HOI) detection has emerged as a key technology in the field of computer vision. HOI detection aims to accurately locate humans and objects in images or videos and recognize the corresponding interaction categories to better understand human activities. Specifically, HOI detection takes an image or video as input and outputs a series of triplets ($<$ human, object, interaction $>$). It is widely used in autonomous driving, action recognition, human-computer interaction, social network analysis, emotion recognition, security monitoring, and video surveillance~\cite{gao2018ican}. This paper primarily reviews the research achievements in image-based HOI detection.

\begin{figure}[!t]
\centering
\includegraphics[width=\linewidth]{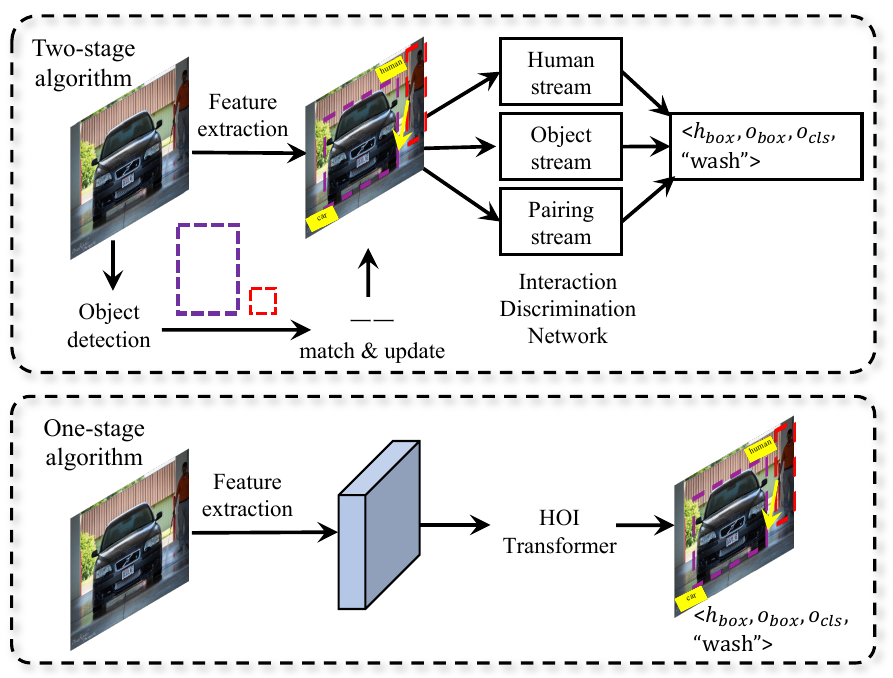}
\caption{%The categories of crowd counting algorithms overviews: (a) Detection-based algorithms estimate the number of people by detecting the head locations.
%(b) Regression-based algorithms directly map image features to global count.
%(c) Density estimation-based methods calculate the headcount by integrating over the crowd density maps.
The flowchart of the HOI detection algorithm.}
\label{fig:figure_1}
\end{figure}

Existing HOI detection algorithms can be roughly divided into two categories: two-stage~\cite{gao2018ican, gkioxari2018detecting, li2019transferable,girshick2015fast, girshick2014rich, chao2018learning, qi2018learning, wan2019pose,gupta2019no, ulutan2020vsgnet, liu2020amplifying, kim2020detecting, zhong2020polysemy,he2021exploiting,li2020detailed, li2020hoi,hou2021affordance, zhang2022efficient} and one-stage (end-to-end)~\cite{liao2020ppdm, zhang2021spatially, fang2021dirv, zhong2021glance, zhang2022exploring, zhou2022human, kim2020uniondet, wang2020learning, zou2021end, dong2021visual, tamura2021qpic, kim2021hotr, chen2021reformulating, ma2023fgahoi, kim2022mstr, peng2023parallel, zhang2021mining, tu2022iwin, kim2023relational, liao2022gen, ning2023hoiclip, wang2024ted, zhang2023exploring,yuan2023rlipv2} methods, as shown in Figure \ref{fig:figure_1}.
The majority of two-stage methods are based on serial models, dividing the HOI task into two steps: human-object detection and interaction classification. Specifically, in the human-object detection stage, a pre-trained object detection framework~\cite{ren2015faster} is typically used to identify all humans and objects present in the image. Then, in the second stage, the detected humans and objects are paired, and their corresponding features are passed to the interaction classification network to determine the type of interaction.
Although two-stage methods are relatively simple to understand, they require iterating over all the information extracted by the detection network for each human-object pair, leading to significant computational overhead. Moreover, while phased training and optimization can enhance the performance of the task at hand, it also results in the loss of contextual information.

Unlike two-stage methods, one-stage approaches can directly output the HOI triplets.
Specifically, one-stage methods usually introduce a new HOI mediator, allowing the network to predict interaction relationships directly.
Compared to two-stage methods, introducing an interaction mediator eliminates the need for separately matching humans and objects, significantly enhancing inference speed. However, one-stage algorithms typically adopt a multi-task collaborative learning architecture, which can lead to interference between tasks.

In real-world scenarios, the variability of human actions and the wide range of object types result in diverse HOI types. Additionally, the high cost of annotation further limits the performance of models. To effectively address these problems, new techniques, including zero-shot learning~\cite{maraghi2021scaling, eum2021semantics, peyre2019detecting}, weakly supervised~\cite{baldassarre2020explanation, kilickaya2021human, wan2023weakly, kumaraswamy2021detecting, zhang2017ppr, wang2024freea, unal2023weakly, zhang2021spatially, peyre2017weakly, sarullo2020zero}, and large-scale language models~\cite{unal2023weakly, shi2022proposalclip, wan2023weakly} are discussed. 

The remainder of this review is organized as follows. In Section \ref{2}, the mainstream datasets
% and evaluation metrics 
are introduced. Section \ref{3} summarizes the milestone HOI
detection technology.
In Section \ref{4},  the strengths and weaknesses of these different methods are summarized, and the future development directions are implemented in the actual architecture. Section \ref{5} explores future development directions.

\section{Datasets}
\label{2}
In the past years, many excellent HOI detection datasets have emerged. Based on annotation granularity, the levels are instance-level, part-level, and pixel-level. The summary and analysis of datasets across these three annotation levels are listed in Table \ref{tab:Dataset Statistics}. Below are their detailed descriptions:

\textbf{V-COCO}\cite{gupta2015visual}: V-COCO is a dataset based on COCO\cite{lin2014microsoft}, containing 10,346 images, with 5,400 used for training and 4,946 for testing. It includes 80 object categories and 29 action categories, 4 of which represent body actions that do not involve any interaction. These interactions cover various actions such as ``eating'', ``reading'', and ``wearing'' associated with objects like ``bread'', ``book'', and ``clothes''.

\textbf{HICO-DET}\cite{chao2018learning}: In 2018, the University of Michigan introduced HICO-DET, a HOI detection dataset with a larger number of images and more complex interaction relationships.
The data is sourced from Flickr and comprises a total of 47,776 images, with 38,118 used for training and 9,658 for testing. The dataset includes 80 object categories and 117 action categories, resulting in 600 different interaction types, of which 462 are non-rare categories and 138 are rare categories. The HICO-DET and V-COCO datasets cover a wide range of object and action categories, which are the most commonly used benchmarks for HOI detection.

\textbf{HCVRD}\cite{zhuang2018hcvrd}: The University of Adelaide in Australia constructed a large-scale human-centered visual relationship detection dataset called HCVRD.
Compared to previously released datasets, this dataset contains a large number of relationship annotations, with nearly 10,000 categories.
This extensive label space more accurately reflects real-world HOIs. 
Unlike HICO-DET and V-COCO, HCVRD not only focuses on interaction relationships but also includes the relative positional relationships between humans and objects.

\textbf{PaStaNet-HOI }\cite{zhuang2018hcvrd}: Under coarse-grained instance-level annotation supervision, the model is prone to overfitting, which leads to poor generalization capabilities. Shanghai Jiao Tong University constructed the PaStaNet-HOI dataset. This dataset provides approximately 110,000 annotated images, with 77,260 images used as the training set, 11,298 images as the validation set, and 22,156 images as the test set. PaStaNet-HOI discards the “no interaction” category. It consists of 116 interaction relationships and 80 object categories, forming a total of 520 HOI relationship categories.

\textbf{PIC }\cite{liu2021human}: Liu et al. construct the pixel-level HOI database PIC (Person in Context) by performing pixel-level annotations on both the human body and objects. This enables more precise localization of humans and objects, even in cases of occlusion. PIC collects 17,122 human-centered images from the internet, which include a training set of 12,339 images, a validation set of 1,916 images, and a test set of 2,867 images.
Additionally, as one of the most finely annotated databases for HOI detection. It includes a comprehensive range of annotation types, such as 141 object categories and 23 interaction relationships between humans and objects.

\begin{table}[hb]
\caption{Dataset Statistics}
\label{tab:Dataset Statistics} 
\centering
\begin{tabular}{c|c|c|c|c}
\bottomrule\noalign{\smallskip}
Dataset      & Total & Object   & Relationship  &Annotation \\ \hline
V-COCO      & 10346   &80        &29   & instance-level\\ 
HICO-DET     & 47776  & 80      & 117  & instance-level\\ 
HCVRD    & 52855    & 1824      & 927  & instance-level\\ 
PaStaNet-HOI  & 110714 & 80     & 116  &part-level\\ 
PIC  & 17122  & 141     & 23 &pixel-level\\ \hline
\toprule
\end{tabular}
\end{table}

% \textbf{Metrics.} 
% At present, there are two main performance evaluation metrics commonly used: mean Average Precision (${A}_{p}$) and mean Accuracy (${m}_{AP}$). The ${A}_{p}$ is defined based on Precision (P) and Recall (r):
% \begin{equation}
%     A_{p} = \sum_{k=1}^{N} P(K) \Delta r (k) \times 100\%,
% \end{equation}
% where N is the total number of test images, $P(k)$ is the precision for recognizing k images, and $ \Delta r (k)$ represents the precision as k varies from $k-1$ to $k$. P is the ratio of positive samples correctly classified to the samples identified as positive after classification, and r is the ratio of positive samples correctly classified to the actual number of positive samples. 

% The ${m}_{AP}$ serves as the mainstream metric for evaluating the detection performance of all categories, which is defined as follows:

% \begin{equation}
%     m_{AP} = \int_{0}^{1} P_1(R)dR,
% \end{equation}
% where $P_1(R)$ represents the normalized recognition accuracy rate.

\section{The architectures of HOI Detection}
\label{3}
Two-stage detection algorithms primarily focus on multi-stream models and graph models. 
One-stage detection algorithms can be divided into bounding box-based, relationship point-based, and query-based models. 
This section briefly reviews the above work and introduces new techniques such as zero-shot learning, weakly supervised, and large-scale language models.

\subsection{Two-stage HOI detection architecture}

Two-stage HOI detection is an instance-guided, bottom-up deep learning approach. Currently, two-stage methods are divided into multi-stream models and graph models. 

\textbf{Multi-stream models.} The multi-stream model is an early attempt in the field of HOI detection. 
Firstly, the object detector generates region proposals about humans and objects. Secondly, an interaction relationship classification network extracts features for the targets and then fuses the classification results. In 2018, Chao et al.~\cite{chao2018learning} construct a widely influential public dataset known as the HICO-A dataset. In addition, they propose a standard two-stage HOI network HO-RCNN for extracting features of spatial relationships between humans and objects. The HO-RCNN is composed of three key components: the human stream, the object stream, and the HOI stream.
Moreover, Gkioxari et al.~\cite{gkioxari2018detecting} propose a new human-centered model called InteractNet, which uses human appearance as a cue to predict the location of target objects. It can simultaneously detect humans, objects, and interaction relationships. 
In work~\cite{li2019transferable}, Li et al utilize an interaction-aware network to learn general interaction knowledge from several HOI datasets. The non-interaction suppression strategy is employed before HOI classification inference, thereby improving network performance.

The subtle visual differences between various interaction relationships bring challenges to HOI detection. 
To address these challenges, Wan et al.~\cite{wan2019pose} propose a multi-level relationship detection strategy that utilizes human pose cues to capture interactions. Specifically, they employ a multi-branch network to learn pose-enhanced relationship representations across three semantic levels. By integrating these features, the model generates robust results.
Zhong et al.~\cite{zhong2020polysemy} propose a new polysemy decoding network, PD-Net, which further mitigates the issue of verb polysemy through three strategies. In work~\cite{iftekhar2023gtnet}, a novel network GTNet based on self-attention guidance, enhances detection performance by encoding spatial context information into the visual features of instances. To address the long-tail distribution problem in HOI,  
Hou et al.~\cite{hou2021detecting} design a novel Fabricated Compositional Learning framework. 
Xu et al.~\cite{xu2022effective} propose a new human-centered framework, with the core idea of using non-local features and human-object part coupling characteristics to detect.

\textbf{Graph models.} The aforementioned multi-stream models overlook the correlations between different human-object pairs. Additionally, processing each human-object pair individually increases time costs. 
To overcome these limitations, Qi et al.~\cite{qi2018learning} first introduce the Graph Parsing Neural Network (GPNN). GPNN uses nodes and edges to identify instances and interaction relationships, respectively. Later, Zhou et al.~\cite{zhou2019relation} propose the Relational Pairwise Neural Network (RPNN), which utilizes object-body part and human-body part relations to analyze pairwise relationships between two graphs.

Previous graph-based algorithms treat humans and objects as the same type of node, failing to distinguish information exchanged between different entities. Therefore, Wang et al.~\cite{wang2020contextual} propose a heterogeneous graph network, CHGNet, which models humans and objects as distinct types of nodes. 
Gao et al.~\cite{gao2020drg} utilize abstract spatial semantic representations to describe each human-object pair. They employ a dual relation graph to aggregate contextual information from the scene and capture discriminative cues, effectively addresses the prediction ambiguity issue.
Most methods primarily focus on the visual and semantic features of instances. However, they do not leverage the high-level semantic relationships within the image. In order to solve this shortcoming, He et al.~\cite{he2021exploiting} embed scene graphs into global contextual cues. Additionally, a message passing module was developed to gather relational information from the neighbors of objects. 
% This work provides essential contextual information and detailed knowledge of interaction relationships for HOI inference.

\subsection{one-stage HOI detection architecture}

One-stage algorithms have surpassed traditional two-stage models in both speed and accuracy. Unlike two-stage methods, one-stage approaches can directly output HOI triplets without an additional object detection process. 

\textbf{Bounding box-based models.} The bounding box-based algorithms directly detect the location and category of targets using a simple structure while simultaneously predicting potential interaction relationships. This simplified design significantly enhances the inference speed of the algorithm.
Previous works first detect instances and then predict interaction actions, which results in longer inference times for HOI detection. To address this challenge, Kim et al.~\cite{kim2020uniondet} propose UnionDet, a method that directly captures interaction regions, eliminating the need for additional inference stages. 
Compared to traditional two-stage algorithms, UnionDet significantly improves inference speed by 4 to 14 times. This innovation makes HOI detection more efficient and real-time.

Traditional one-stage methods typically focus on the joint region of interaction, which can introduce unnecessary visual noises. 
To tackle this issue, Fang et al.~\cite{fang2021dirv} propose DIRV, which focuses on the interaction regions of each human-object pair at different scales and extracts the most relevant subtle visual features. 
Additionally, DIRV develops a voting strategy that leverages the overlapping parts within the interaction region, replacing the traditional non-maximum suppression method.

\textbf{Relationship point-based models.} Inspired by anchor-free detectors, research based on relationship points opens a new era in one-stage methods.
Wang et al.~\cite{wang2020learning} argue that extracting appearance features only is insufficient to handle complex HOI sciences.
Therefore, they propose IP-Net, the first algorithm to view HOI detection as a keypoint detection problem.
PPDM~\cite{liao2020ppdm} is the first real-time HOI detection method that redefines the HOI triplet as $<$human point, interaction point, object point$>$. As a novel parallel architecture, PPDM significantly reduces computational costs by filtering of interaction points.

Existing one-stage models lack a reasoning step for dynamic discriminative cues. Zhong et al. further improve PPDM by GGNet~\cite{zhong2021glance}. GGNet first determines whether a pixel is an interaction point, then infers action-aware points around the pixel to refine the point's position. 

\begin{table*}[!ht]
\centering
\caption{Performance comparisons on HICO-Det dataset. * represents the results given in~\cite{wan2023weakly,unal2023weakly}.}
\label{tab-hico-det}
\begin{tabular}{lll|ccc|ccc}
\hline
\multirow{2}{*}{Methods}                                                 & \multirow{2}{*}{Backbone} & \multicolumn{1}{l|}{\multirow{2}{*}{Source}} & \multicolumn{3}{c|}{Default (\textbf{mAP}$\uparrow$)}                                                         & \multicolumn{3}{c}{Know Object (\textbf{mAP}$\uparrow$)}                                                     \\
                                                                         &                           & \multicolumn{1}{l|}{}                          & \multicolumn{1}{l}{Full} & \multicolumn{1}{l}{Rare} & \multicolumn{1}{l|}{None-Rare} & \multicolumn{1}{l}{Full} & \multicolumn{1}{l}{Rare} & \multicolumn{1}{l}{None-Rare} \\ \hline
% \multirow{3}{*}{Method}                                                         & \multirow{3}{*}{Backbone} & \multicolumn{1}{c|}{\multirow{3}{*}{Source}} & \multicolumn{6}{c|}{HICO-Det}                                                               & \multicolumn{2}{c}{V-COCO}                                       \\ \cline{4-11} 
%                                                                                 &                           & \multicolumn{1}{c|}{}                        & \multicolumn{3}{c|}{Default}                 & \multicolumn{3}{c|}{Know Object}             & \multicolumn{1}{c|}{\multirow{2}{*}{AP$^{S1}_{role}$}} & \multirow{2}{*}{AP$^{S1}_{role}$} \\ \cline{4-9}
%                                                                                 &                           & \multicolumn{1}{c|}{}                        & Full & Rare & \multicolumn{1}{c|}{None-Rare} & Full & Rare & \multicolumn{1}{c|}{None-Rare} & \multicolumn{1}{c|}{}                     &                      \\ \hline

\multicolumn{9}{c}{Two-Stage Methods}                                                                                                                                                                                                                                                                                                \\
HO-RCNN~\cite{chao2018learning}           & CaffeNet             & \multicolumn{1}{l|}{WACV 2018}                 & 7.81                     & 5.37                     & \multicolumn{1}{c|}{8.54}     & 10.41                        & 8.94                        & 10.85                         \\
InteractNet~\cite{gkioxari2018detecting}           & ResNet-50-FPN             & \multicolumn{1}{l|}{CVPR 2018}                 & 9.94                     & 7.16                     & \multicolumn{1}{c|}{10.77}     & -                        & -                        & -                         \\
GPNN~\cite{qi2018learning}           & ResNet-101             & \multicolumn{1}{l|}{ECCV 2018}                 & 13.11                     & 9.34                     & \multicolumn{1}{c|}{14.23}     & -                        & -                        & -                          \\
iCAN~\cite{gao2018ican}                            & ResNet-50                 & \multicolumn{1}{l|}{BMCV 2018}                 & 14.84                    & 10.45                    & \multicolumn{1}{c|}{16.15}     & 16.26                    & 11.33                    & 17.73                         \\
PMFNet~\cite{wan2019pose}                          & ResNet-50-FPN             & \multicolumn{1}{l|}{ICCV 2019}                 & 17.46                    & 15.56                    & \multicolumn{1}{c|}{18.00}     & 20.34                    & 17.47                    & 21.20                         \\
No-Frills~\cite{gupta2019no}                          & ResNet-152             & \multicolumn{1}{l|}{ICCV 2019}                 & 17.18                     & 12.17                    & \multicolumn{1}{c|}{18.68}     & -                    & -                    & -                         \\
VSGNet~\cite{ulutan2020vsgnet}                          & ResNet-152             & \multicolumn{1}{l|}{CVPR2020}                 & 19.80                       & 16.05                   & \multicolumn{1}{c|}{20.91}     & -                    & -                    & -                           \\
FCMNet~\cite{liu2020amplifying}                          & ResNet-50             & \multicolumn{1}{l|}{ECCV2020}                 & 20.41                       & 17.34                   & \multicolumn{1}{c|}{21.56}     & 22.04                    & 18.97                    & 23.12                   \\

ACP~\cite{kim2020detecting}              & Res-DCN-152     & ECCV 2020    & 20.59   & 15.92  & 21.98     & -        & -       & -          \\
PD-Net~\cite{zhong2020polysemy}          & ResNet-152-FPN     & ECCV 2020 & 20.81   & 15.90  & 22.28     & 24.78    & 18.88   & 26.54      \\
SG2HOI~\cite{he2021exploiting}           & ResNet-50     & ICCV 2021      & 20.93   & 18.24  & 21.78     & 24.83    & 20.52   & 25.32      \\
DJ-RN~\cite{li2020detailed}              & ResNet-50     & CVPR 2020      & 21.34   & 18.53  & 22.18     & 23.69    & 20.64   & 24.60      \\
SCG~\cite{zhang2021spatially}            & ResNet-50-FPN     & ICCV 2021  & 21.85   & 18.11  & 22.97     & -        & -       & -          \\
IDN~\cite{li2020hoi}                     & ResNet-50     & NIPS 2020      & 23.36   & 22.47  & 23.63     & 26.43    & 25.01   & 26.85      \\
ATL~\cite{hou2021affordance}             & ResNet-50 & CVPR 2021      & 23.81   & 17.43  & 25.72     & 27.38    & 22.09   & 28.96      \\

% DJ-RN~\cite{li2020detailed}                        & ResNet-50                 & \multicolumn{1}{l|}{CVPR 2020}                 & 21.34                    & 18.53                    & \multicolumn{1}{c|}{22.18}     & 23.69                    & 20.64                    & 24.60                         \\

% IDN~\cite{li2020hoi}                               & ResNet-50                 & \multicolumn{1}{l|}{NeurIPS 2020}              & 23.36                    & 22.47                    & \multicolumn{1}{c|}{23.63}     & 26.43                    & 25.01                    & 26.85                         \\

UPT~\cite{zhang2022efficient} &  ResNet-101-DC5 & CVPR 2022 & \textbf{32.62} & \textbf{28.62} & \textbf{33.81} & \textbf{36.08} & \textbf{31.41} & \textbf{37.47} \\

\hline
\multicolumn{9}{c}{One-Stage Methods}                                                                                                                                                                                                                                                                                             \\
UnionDet~\cite{kim2020uniondet}          & ResNet-50-FPN     & ECCV 2020       & 17.58   & 11.72  & 19.33     & 19.76    & 14.68   & 21.27      \\
IP-Net~\cite{wang2020learning}           & Hourglass-104     & CVPR 2020      & 19.56   & 12.79  & 21.58     & 22.05    & 15.77   & 23.92      \\
HRNet~\cite{tip-9552553}                     & ResNet-152                 & \multicolumn{1}{l|}{TIP 2021}                 & 21.93                    & 16.30                    & \multicolumn{1}{c|}{23.62}     & 25.22                    & 18.75                    & 27.15                         \\
PPDM-Hourglass~\cite{liao2020ppdm}       & Hourglass-104 & CVPR 2020       & 21.94   & 13.97  & 24.32     & 24.81    & 17.09   & 27.12      \\
HOI-Trans~\cite{zou2021end}              & ResNet-50 & CVPR 2021           & 23.46   & 16.91  & 25.41     & 26.15    & 19.24   & 28.22      \\
GG-Net~\cite{zhong2021glance}            & Hourglass-104 & CVPR 2021       & 23.47   & 16.48  & 25.60     & 27.36    & 20.23   & 29.48      \\
PST~\cite{dong2021visual}                & ResNet-50        & ICCV 2021           & 23.93   & 14.98  & 26.60     & 26.42    & 17.61   & 29.05      \\
HOTR~\cite{kim2021hotr}                  & ResNet-50 & CVPR 2021           & 25.10   & 17.34  & 27.42     & -        & -       & -          \\
AS-Net~\cite{chen2021reformulating}      & ResNet-50 & CVPR 2021           & 28.87   & 24.25  & 30.25     & 31.74    & 27.07   & 33.14      \\
QPIC~\cite{tamura2021qpic}               & ResNet-50 & CVPR 2021          & 29.07   & 21.85  & 31.23     & 31.68    & 24.14   & 33.93      \\
FGAHOI~\cite{ma2023fgahoi}               & Swin-Tiny & TPAMI 2023           & 29.94   & 22.24  & 32.24     & 32.48    & 24.16   & 34.97      \\
MSTR~\cite{kim2022mstr}                            & ResNet-50                 & \multicolumn{1}{l|}{CVPR 2022}                 & 31.17                    & 25.31                    & \multicolumn{1}{c|}{33.92}     & 34.02                    & 28.83                    & 35.57                         \\
PR-Net~\cite{peng2023parallel}               & ResNet-50 & arXiv 2023           & 31.17   & 25.66  & 32.82     & -    & -   & -      \\
CDN-S~\cite{zhang2021mining}               & ResNet-50 & NIPS 2021           & 31.44   & 27.39  & 32.64     & 34.09    & 29.63   & 35.42      \\
DisTr~\cite{zhou2022human}                         & ResNet-50                 & \multicolumn{1}{l|}{CVPR 2022}                 & 31.75                    & 27.45                    & \multicolumn{1}{c|}{33.03}     & 34.50                    & 30.13                    & 35.81                         \\
CDN-B~\cite{zhang2021mining}               & ResNet-50 & NIPS 2021           & 31.78   & 27.55  & 33.05     & 34.53    & 29.73   & 35.96      \\
Iwin-B~\cite{tu2022iwin}               & ResNet-50-FPN & ECCV 2022          & 32.03   & 27.62  & 34.14     & 35.17    & 28.79   & 35.91      \\
RCL~\cite{kim2023relational}                       & ResNet-50                 & \multicolumn{1}{l|}{CVPR 2023}                 & 32.87                    & 28.67                    & \multicolumn{1}{c|}{34.12}     & 35.52                    & 30.88                    & 36.45                         \\
GEN-VLKT~\cite{liao2022gen}              & ResNet-50 & CVPR 2022           & 33.75   & 29.25  & 35.10     & 36.78    & 32.75   & 37.99      \\
% HOTR~\cite{kim2021hotr}                            & ResNet-50                 & \multicolumn{1}{l|}{CVPR 2021}                 & 25.10                    & 17.34                    & \multicolumn{1}{c|}{27.42}     & -                        & -                        & -                             \\
% QPIC~\cite{tamura2021qpic}                         & ResNet-101                & \multicolumn{1}{l|}{CVPR 2021}                 & 29.90                    & 23.92                    & \multicolumn{1}{c|}{31.69}     & 32.38                    & 26.06                    & 34.27                         \\

% GEN-VLKT~\cite{liao2022gen}                        & ResNet-50                 & \multicolumn{1}{l|}{CVPR 2022}                 & 33.75                    & 29.25                    & \multicolumn{1}{c|}{35.10}     & 36.78                    & 32.75                    & 37.99                         \\
HOICLIP~\cite{ning2023hoiclip}                     & ResNet-50                 & \multicolumn{1}{l|}{CVPR 2023}                 & 34.59                    & 31.12                    & \multicolumn{1}{c|}{35.74}     & 37.61                    & 34.47                    & 38.54                         \\

PBLQG~\cite{tip-10328553}                     & ResNet-50                 & \multicolumn{1}{l|}{TIP 2023}                 & 31.64                    & 26.23                    & \multicolumn{1}{c|}{33.25}     & 34.61                    & 30.16                    & 35.93                         \\
SG2HOI~\cite{tip-10315051}                     & ResNet-50                 & \multicolumn{1}{l|}{TIP 2023}                 & 33.14                    & 29.27                    & \multicolumn{1}{c|}{35.72}     & 35.73                    & 32.01                    & 36.43                         \\
TED-Net~\cite{wang2024ted}                     & ResNet-50                 & \multicolumn{1}{l|}{TCSVT 2024}                 & 34.00                    & 29.88                    & \multicolumn{1}{c|}{35.24}     & 37.13                    & 33.63                    & 38.18 \\      
PViC~\cite{zhang2023exploring}                     & ResNet-50                 & \multicolumn{1}{l|}{ICCV 2023}                 & 34.69                    & 32.14                    & 35.45 &  38.14 & 35.38 & 38.97        \\ 
PViC~\cite{zhang2023exploring}                     & Swin-L                 & \multicolumn{1}{l|}{ICCV 2023}                 & \textbf{44.32}                    & 44.61                    & 44.24 & \textbf{47.81} & \textbf{48.38} & \textbf{47.64}        \\
RLIPv2~\cite{yuan2023rlipv2}                     & Swin-L                 & \multicolumn{1}{l|}{ICCV 2023}                 & 43.23                    & \textbf{45.64}                    & \textbf{45.09} & - & - & -      
\\\hline

\multicolumn{9}{c}{Weakly+ Supervised Methods}                                                                                                                                                                                                                                                                                             \\
Explanation-HOI*~\cite{baldassarre2020explanation} & ResNeXt101                & \multicolumn{1}{l|}{ECCV 2020}                 & 10.63                    & 8.71                     & \multicolumn{1}{c|}{11.20}     & -                        & -                        & -                             \\
MAX-HOI~\cite{kumaraswamy2021detecting}            & ResNet101                 & \multicolumn{1}{l|}{WACV 2021}                 & 16.14                    & 12.06                    & \multicolumn{1}{c|}{17.50}     & -                        & -                        & -                             \\
Align-Former~\cite{kilickaya2021human}              & ResNet-101                & \multicolumn{1}{l|}{arXiv 2021}                & 20.85                    & 18.23                    & \multicolumn{1}{c|}{21.64}     & -                        & -                        & -                             \\
PPR-FCN*~\cite{zhang2017ppr}                       & ResNet-50                 & \multicolumn{1}{l|}{ICCV 2017}                 & 17.55                    & 15.69                    & \multicolumn{1}{c|}{18.41}     & -                        & -                        & -                             \\
% Weakly-HOI~\cite{unal2023weakly}                   & ResNet-50                 & \multicolumn{1}{l|}{CVPR 2023}                & 19.26                    & -                        & \multicolumn{1}{c|}{-}         & -                        & -                        & -                             \\
PGBL~\cite{wan2023weakly}                          & ResNet-50                 & \multicolumn{1}{l|}{ICLR 2023}                 & 22.89                    &\textbf{22.41}                    & \multicolumn{1}{c|}{23.03}     & -                        & -                        & -                             \\
FreeA~\cite{wang2024freea}                                                                     & ResNet-50                 & arXiv 2024                          & \textbf{24.57}                    & 21.45                    & \multicolumn{1}{c|}{\textbf{25.51}}     &       \textbf{26.52   }&   \textbf{23.64}                       &          \textbf{27.38}                     \\ \hline
\multicolumn{9}{c}{Weakly Supervised Methods}                                                                                                                                                                                                                                                                                              \\
SCG*~\cite{zhang2021spatially}                     & ResNet-50                 & \multicolumn{1}{l|}{ICCV 2021}                 & 7.05                     & -                        & \multicolumn{1}{c|}{-}         & -                        & -                        & -                             \\
VLHOI~\cite{unal2023weakly}                                                               & ResNet-50                 & \multicolumn{1}{l|}{CVPR 2023}                & 8.38                     & -                        & \multicolumn{1}{c|}{-}         & -                        & -                        & -                             \\ 
% \multicolumn{9}{c}{Weakly\text{-}\text{-} supervised}                                                                                                                                                                                                                                                                                            \\
FreeA~\cite{wang2024freea}                                                                     & ResNet-50                 & \multicolumn{1}{c|}{arXiv 2024}                         & \textbf{16.96}                    & \textbf{16.26}                          & \multicolumn{1}{c|}{\textbf{17.17}}          &      \textbf{18.89}                    &    \textbf{18.11}                      &      \textbf{19.12}                         \\ \hline
\end{tabular}
\end{table*}

\textbf{Query-based models.} Tamura et al.~\cite{tamura2021qpic} propose a transformer-based feature extractor, where the attention mechanism and query-based detection play key roles. One-stage HOI detection algorithms based on the transformer architecture are gradually emerging and developing. Unlike existing transformer-based models that query at a single level, Dong et al.~\cite{dong2021visual} explicitly merge and sum queries to better model the relationships between parts and the whole, which are not directly captured within the transformer. Kim et al.~\cite{kim2020uniondet} use HOTR to predict triplets from images directly. This method effectively leverages the inherent semantic relationships within the image, eliminating the post-processing strategies used in previous approaches.
Moreover, Liao et al.~\cite{liao2022gen} propose a dual-branch GEN-VLKT, which eliminates the need for post-matching. CLIP~\cite{radford2021learning} is also embedded to initialize the classifier, effectively leveraging image and text information. Similarly, the category-aware transformer network (CATN)~\cite{dong2022category} enhances detector performance by initializing object queries with category-aware semantic information. In work~\cite{lim2023ernet}, ERNet utilizes multi-scale deformable attention to capture essential HOI features.

\subsection{New techniques} This section introduces new techniques, including zero-shot learning, weakly supervised, and large-scale language models.

\textbf{Zero-shot learning.}
Maraghi et al.~\cite{maraghi2021scaling} is the first to utilize zero-shot learning methods to address the long-tail problem in HOI detection. In this work, a decomposed model is used to separately infer verbs and objects, enabling the detection of new verb-object combinations during the testing phase.
Eum et al.~\cite{eum2021semantics} propose an HOI detection method based on verb-object relationship reasoning, where semantic and spatial information is embedded into the visual stream. 
Due to the combinatorial nature of visual relationships, collecting a sufficiently large amount of trainable triplet data is hard. To address this, Peyre et al.~\cite{peyre2019detecting} develop a model that successfully merges semantic and visual spaces.

\textbf{Weakly supervised models.}
Most existing HOI detection models require fully annotated data for supervised training. However, obtaining complete data labels remains a challenging task. To effectively address this issue, weakly supervised HOI detection typically uses image-level interaction labels for training. For example, Peyre et al.~\cite{peyre2017weakly} introduce a weakly supervised discriminative clustering model that learns relationships solely from image-level labels. They also propose a new and challenging dataset, UnRel, to accurately evaluate visual relationships.
In work~\cite{sarullo2020zero}, Sarullo et al. model the relationships between actions and objects in the form of a graph.
Align-Former~\cite{kilickaya2021human} is equipped with an HOI alignment layer that generates pseudo aligned human-object pairs based on weakly supervised. 
Furthermore, Baldassarre et al.~\cite{baldassarre2020explanation} attempt to leverage a multi instance learning framework to detect instances and then use image-level labels to supervise the interaction classifier. 

\textbf{Large-scale language models.} With the help of large language models, Unal et al.~\cite{unal2023weakly} use only image-level labels to query possible interactions between human and object categories. The graph-based ProposalCLIP~\cite{shi2022proposalclip} addresses the limitations of CLIP cues by predicting the categories of objects without annotations, effectively enhancing performance. 
Recently, Wan et al.~\cite{wan2023weakly} develop a CLIP-guided HOI representation that integrates prior knowledge at both the image level and the human-object pair level. This dual-layer framework is designed to more effectively utilize image-level information through a shared HOI knowledge base, thereby further enhancing the learning of interaction features.

\section{Complex problem of HOI detection}
\label{4}
Table \ref{tab-hico-det} and Table \ref{tab-vcoco} present the results of different methods on the HICO-DET and V-COCO datasets, respectively.
The advantage of two-stage methods is they can decouple object detection and interaction classification, allowing each stage to focus on optimizing its specific task. 
However, this approach encounters many obstacles.
There are imbalanced positive and negative sample distributions, additional computational, and insufficient information exchange.

One-stage HOI detection algorithms significantly improve efficiency and accuracy by directly predicting interactions. However, these models typically use a multi-task learning approach to share features, which may lead to interference between tasks, preventing the model from achieving optimal performance.

Significant progress has been made in the field of HOI detection, especially with the emergence of new technologies. These methods can directly predict interactions, avoiding the complexity of the two-stage process. Nevertheless, despite these advancements, several issues remain in the field of HOI detection. Current models may perform poorly when handling multiple interaction types. Additionally, environmental conditions and lighting issues can further affect the detection of small objects.

\begin{table}[h]
\centering
\caption{Performance comparisons on V-COCO dataset.}
\label{tab-vcoco}
\begin{tabular}{llcc}
\hline
Method                      & Source   & AP1  & AP2   \\
           &   & (\textbf{mAP$\uparrow$})   &  (\textbf{mAP$\uparrow$})  \\ \hline
\multicolumn{4}{c}{Two-stage Methods}\\
                                             % &          &       &       \\
InteractNet~\cite{gkioxari2018detecting} & CVPR 2018 & 40.0  & -     \\
GPNN~\cite{qi2018learning}               & ECCV 2018 & 44.0  & -     \\
iCAN~\cite{gao2018ican}                  & BMCV 2018 & 45.3  & 52.4  \\
TIN~\cite{li2019transferable}            & CVPR 2019 & 47.8  & 54.2  \\
VCL~\cite{hou2020visual}                 & ECCV 2020 & 48.3  & -     \\
DRG~\cite{gao2020drg}                    & ECCV 2020 & 51.0  & -     \\
IP-Net~\cite{wang2020learning}           & CVPR 2020 & 51.0  & -     \\
VSGNet~\cite{ulutan2020vsgnet}           & CVPR 2020 & 51.8  & 57.0  \\
PMFNet~\cite{wan2019pose}                & ICCV 2019 & 52.0  & -     \\
PD-Net~\cite{zhong2020polysemy}          & ECCV 2020 & 52.6  & -     \\
FCMNet~\cite{liu2020amplifying}          & ECCV 2020 & 53.1  & -     \\
ACP~\cite{kim2020detecting}              & ECCV 2020 & 53.2 & -     \\
IDN~\cite{li2020hoi}                     & NIPS 2020 & 53.3  & 60.3  \\ 
UPT~\cite{zhang2022efficient}            & CVPR 2022 & \textbf{61.3}  & \textbf{67.1}  \\ \hline
\multicolumn{4}{c}{One-stage Methods}\\
% :                                             &          &       &       \\
UnionDet~\cite{kim2020uniondet}          & ECCV 2020      & 47.5  & 56.2  \\
HOI-Trans~\cite{zou2021end}              & CVPR 2021      & 52.9  & -     \\
AS-Net~\cite{chen2021reformulating}      & CVPR 2021      & 53.9  & -     \\
GG-Net~\cite{zhong2021glance}            & CVPR 2021      & 54.7  & -     \\
HOTR~\cite{kim2021hotr}                  & CVPR 2021      & 55.2  & 64.4  \\
QPIC~\cite{tamura2021qpic}               & CVPR 2021      & 58.8  & 61.0  \\
Iwin-B~\cite{tu2022iwin}               & ECCV 2022      & 60.5  & -  \\
FGAHOI~\cite{ma2023fgahoi}               & TPAMI 2023      & 60.5  & 61.2  \\
Iwin-L~\cite{tu2022iwin}               & ECCV 2022      & 60.9  & -  \\
PR-Net~\cite{peng2023parallel}               & arXiv 2023      & 61.4  & -  \\
CDN-S~\cite{zhang2021mining}               & NIPS 2021      & 61.7  & 63.8  \\
CND-B~\cite{zhang2021mining}               & NIPS 2021      & 62.3  & 64.4  \\
GEN-VLKT~\cite{liao2022gen}              & CVPR 2022      & 62.4 & 64.5 \\
TED-Net~\cite{wang2024ted}                                  & TCSVT 2024      & 63.4 & 65.0    \\
PViC~\cite{zhang2023exploring} & ICCV 2023     & 64.1 & 70.2    \\
RLIPv2~\cite{yuan2023rlipv2} & ICCV 2023     & \textbf{72.1} & \textbf{74.1}    \\
\hline
\multicolumn{4}{c}{Weakly+ Supervised Methods}\\
Align-Former~\cite{kilickaya2021human}               & arXiv 2021            & \multicolumn{1}{c}{15.8}                                                                & \multicolumn{1}{c}{16.3}                                                                \\
PGBL~\cite{wan2023weakly}                           & ICLR 2023            & \multicolumn{1}{c}{43.0}                                                                & \multicolumn{1}{c}{48.1}                                                                \\
FreeA~\cite{wang2024freea}                                                                      &   \multicolumn{1}{c}{arXiv 2024}                    & \multicolumn{1}{c}{\textbf{50.2}}                                                                & \multicolumn{1}{c}{\textbf{52.1}}                                                               \\ \hline
\multicolumn{4}{c}{Weakly Supervised Methods}\\
SCG*~\cite{zhang2021spatially}        & \multicolumn{1}{l}{ICCV 2021}             & 20.1 & - \\
VLHOI†~\cite{unal2023weakly}       & \multicolumn{1}{l}{CVPR 2023}              & 17.7 & - \\
VLHOI~\cite{unal2023weakly}       & \multicolumn{1}{l}{CVPR 2023} & 29.6 & - \\
FreeA~\cite{wang2024freea}                                                                      &   \multicolumn{1}{c}{arXiv 2024}                    & \multicolumn{1}{c}{\textbf{30.8}}                                                                & \multicolumn{1}{c}{\textbf{32.6}}            \\        
\hline
\end{tabular}
\end{table}

Peering into the future, solutions to these issues could begin with improving the diversity and quality of datasets. Techniques such as transfer learning can also be utilized to enhance the generalizability of models. More effective multi-task learning strategies can be explored to better balance the relationships between different tasks and improve overall performance. Moreover, incorporating more contextual information, such as semantic or temporal data, could further enhance the accuracy of interaction detection.

\section{Conclusion}
\label{5}
HOI detection is a research hotspot in computer vision, with widespread applications in action recognition, pose estimation, autonomous driving, and other scenarios. This paper discusses the latest advancements in two-stage and one-stage HOI detection tasks, providing an overview of the cutting-edge developments. At the same time, new techniques including zero-shot learning, weakly supervised learning, and large-scale language methods are introduced. Finally, the current challenges faced by HOI detection, from the perspective of technical difficulties, are summarized, and its future development trends are predicted.

\bibliographystyle{IEEEtran}
\bibliography{ref}

% \begin{thebibliography}{00}
% \bibitem{b1} G. Eason, B. Noble, and I. N. Sneddon, ``On certain integrals of Lipschitz-Hankel type involving products of Bessel functions,'' Phil. Trans. Roy. Soc. London, vol. A247, pp. 529--551, April 1955.
% @article{廖越2022基于深度学习的人,
%   title={基于深度学习的人—物交互关系检测综述},
%   author={廖越 and 李智敏 and 刘偲},
%   year={2022},
%   publisher={中国图象图形学报}
% }
% \bibitem{b2} J. Clerk Maxwell, A Treatise on Electricity and Magnetism, 3rd ed., vol. 2. Oxford: Clarendon, 1892, pp.68--73.
% \bibitem{b3} I. S. Jacobs and C. P. Bean, ``Fine particles, thin films and exchange anisotropy,'' in Magnetism, vol. III, G. T. Rado and H. Suhl, Eds. New York: Academic, 1963, pp. 271--350.
% \bibitem{b4} K. Elissa, ``Title of paper if known,'' unpublished.
% \bibitem{b5} R. Nicole, ``Title of paper with only first word capitalized,'' J. Name Stand. Abbrev., in press.
% \bibitem{b6} Y. Yorozu, M. Hirano, K. Oka, and Y. Tagawa, ``Electron spectroscopy studies on magneto-optical media and plastic substrate interface,'' IEEE Transl. J. Magn. Japan, vol. 2, pp. 740--741, August 1987 [Digests 9th Annual Conf. Magnetics Japan, p. 301, 1982].
% \bibitem{b7} M. Young, The Technical Writer's Handbook. Mill Valley, CA: University Science, 1989.
% \end{thebibliography}
\vspace{12pt}
% \color{red}
% IEEE conference templates contain guidance text for composing and formatting conference papers. Please ensure that all template text is removed from your conference paper prior to submission to the conference. Failure to remove the template text from your paper may result in your paper not being published.

\end{document}